%% file: ijcai24.tex
\documentclass{article}
\usepackage{ijcai24}

\usepackage{times}
\usepackage{soul}
\usepackage{url}
\usepackage[hidelinks]{hyperref}
\usepackage[utf8]{inputenc}
\usepackage[small]{caption}
\usepackage{graphicx}
\usepackage{amsmath}
\usepackage{amsthm}
\usepackage{booktabs}
\usepackage{algorithm}
\usepackage{algorithmic}
\usepackage[switch]{lineno}
\input{math_commands.tex}

\usepackage{hyperref}
\usepackage{url}

\usepackage{caption}
\usepackage{subcaption}
\usepackage{graphicx}

\usepackage{multirow}
\usepackage{booktabs}
\usepackage{subcaption}
\usepackage{nicematrix,tikz,makecell}
\usepackage{wrapfig}
\usepackage{amsthm}
\usepackage{amsmath}
\usepackage{amsthm}
\usepackage{amsfonts}
\usepackage{bbm}
\usepackage{bm}
\usepackage{mathtools}

\usepackage{xspace}
\newcommand{\our}{TrainlessGNN\xspace}

\usepackage{xcolor}        
\newcommand{\firstt}[2]{\textcolor{red}{$\mathbf{{#1} {\scriptstyle \pm {#2}}}$}}%
\newcommand{\secondt}[2]{\textcolor{blue}{$\mathbf{{#1} {\scriptstyle \pm {#2}}}$}}%
\newcommand{\thirdt}[2]{\textcolor{violet}{$\mathbf{{#1} {\scriptstyle \pm {#2}}}$}}%

\definecolor{MyRed}{HTML}{EA6B66}
\definecolor{MyGreen}{HTML}{97D077}

\newtheorem{assumption}{Assumption}

\title{You do not have to train Graph Neural Networks at all on text-attributed graphs}

\author{
Kaiwen Dong$^1$
\and
Zhichun Guo$^1$\And
Nitesh V. Chawla$^1$\\
\affiliations
$^1$University of Notre Dame\\
\emails
\{kdong2, zguo5, nchawla\}@nd.edu
}

\begin{document}

\maketitle

\begin{abstract}
Graph structured data, specifically text-attributed graphs (TAG), effectively represent relationships among varied entities. Such graphs are essential for semi-supervised node classification tasks. Graph Neural Networks (GNNs) have emerged as a powerful tool for handling this graph-structured data. Although gradient descent is commonly utilized for training GNNs for node classification, this study ventures into alternative methods, eliminating the iterative optimization processes. We introduce TrainlessGNN, a linear GNN model capitalizing on the observation that text encodings from the same class often cluster together in a linear subspace. This model constructs a weight matrix to represent each class's node attribute subspace, offering an efficient approach to semi-supervised node classification on TAG. Extensive experiments reveal that our trainless models can either match or even surpass their conventionally trained counterparts, demonstrating the possibility of refraining from gradient descent in certain configurations.
\end{abstract}
\vspace{-2mm}
\input{body/introduction}
\vspace{-2mm}
\input{body/method}
\vspace{-2mm}
\input{body/experiment}
\vspace{-2mm}
\input{body/conclusion}


\bibliographystyle{named}
\bibliography{references}

\newpage
\appendix
\onecolumn
\input{body/appendix}

\end{document}

%% file: math_commands.tex
\usepackage{amsmath,amsfonts,bm}

\def\Figref#1{Figure~\ref{#1}}

\def\eqref#1{equation~\ref{#1}}
\def\Eqref#1{Equation~\ref{#1}}

\def\1{\bm{1}}

\DeclareMathAlphabet{\mathsfit}{\encodingdefault}{\sfdefault}{m}{sl}
\SetMathAlphabet{\mathsfit}{bold}{\encodingdefault}{\sfdefault}{bx}{n}



%% file: body/introduction.tex
\section{Introduction}
Graph structured data is widely used across many fields due to its ability to show relationships between different entities~\cite{wu_comprehensive_2021}. In many cases, the nodes in these graphs are associated with text attributes, leading to what's known as text-attributed graphs (TAG)~\cite{chen_exploring_2023}. TAG has versatile applications, including social media~\cite{liben-nowell_link_2003}, citation networks~\cite{yang_revisiting_2016}, academic collaborations~\cite{hu_open_2021}, or recommendation system~\cite{shchur_pitfalls_2019}.


We aim to delve into effective approaches for handling TAG, focusing specifically on semi-supervised node classification tasks~\cite{yang_revisiting_2016}. The objective here is to predict the labels of unlabeled nodes within a graph, utilizing a limited set of labeled nodes as a reference~\cite{kingma_semi-supervised_2014}. To effectively capture both the node attributes and the topological structure within the graph, Graph Neural Networks (GNNs)~\cite{kipf_semi-supervised_2017,hamilton_inductive_2018,velickovic_graph_2018}, especially those of the message-passing type~\cite{gilmer_neural_2017}, have demonstrated significant success in effectively managing graph-structured data. Typically, the GNN process begins by transforming the textual data of each node into a vector using text embedding techniques like Bag-Of-Words (BOW), TF-IDF, and word2vec~\cite{mikolov_distributed_2013}\footnote{Recently, there has been growing interest in utilizing Large Language Models to encode textual data. However, employing a more complex text encoder alone doesn't offer substantial benefits over simpler embeddings like BOW and TF-IDF~\cite{purchase_revisiting_2022,chen_exploring_2023}.}. The straightforward application of these shallow text embedding methods has led to their widespread adoption as the go-to text encoding technique in numerous graph benchmark datasets~\cite{yang_revisiting_2016,hu_open_2021,shchur_pitfalls_2019} (see Table \ref{tab:stats}).

When it comes to node classification with the textual data encoded, a GNN is trained on the graph to fit the data accurately. The training phase, often seen as an optimization process, commonly employs iterative tools like gradient descent to update the model's weight to minimize a predefined loss function over the training samples. Although gradient descent (and its variants~\cite{sun_survey_2019}) has become almost synonymous with model fitting, it raises a question whether there are alternative methods to fit the GNNs.

One such alternative is proposed by UGT \cite{huang_you_2022}, drawing inspiration from the lottery ticket hypothesis \cite{zhou_deconstructing_2019,frankle_lottery_2018}. UGT suggests fitting GNNs without updating model weights by identifying a mask to sparsify untrained neural networks. Although these sparse subnetworks can perform comparably to trained dense networks, finding an appropriate mask involves an iterative discrete optimization process, which can be even more computationally demanding than traditional gradient descent optimization.


In this paper, we explore how \textbf{GNNs can be fitted without employing traditional iterative processes like gradient descent to tackle semi-supervised node classification tasks on TAG}. Given the absence of a closed-form solution for multiclass classification problems,  we suggest approximating optimal parameters by harnessing both the node attributes and graph structures. We observe that on TAG, text encodings from the same class tend to cluster in the same linear subspace, while being orthogonal to those from different classes. Moreover, we investigate the training dynamics of GCN~\cite{kipf_semi-supervised_2017} and SGC~\cite{wu_simplifying_2019}, two representative GNNs. Our findings suggest that traditional GNN training on TAG can be seen as a process to locate the weight vectors close to the text encodings from corresponding classes in the linear space. Inspired by these insights, we introduce \our, a method that fits a linear GNN model by constructing a weight matrix reflecting the subspace of a particular class's node attributes. The formulation of the weight matrix in our approach can be interpreted as a closed-form solution for a linear regression problem, solved by minimum-norm interpolation in an over-parameterized regime~\cite{wang_benign_2023}, offering a novel pathway for addressing semi-supervised node classification on TAG.




To summarize, our contributions are as follows:
\begin{itemize}
\item We investigate the training dynamics of common GNNs like GCN and SGC on TAG. We discover that the weight matrix is fundamentally pushed towards approximating the subspaces of the node attributes associated with respective classes.
\item We introduce \our, an innovative and efficient method for semi-supervised node classification. To our knowledge, \our is the first to achieve significant predictive performance without the need for an iterative training process in fitting GNN models.
\item Through empirical evaluation on various TAG benchmarks, we demonstrate that our method, devoid of a typical training process, can either match or surpass the performance of conventionally trained models.
\end{itemize}

%% file: body/method.tex
\section{Preliminaries and Related Work}
\paragraph{Notations.}
We examine a graph $ \mathcal{G} = (\mathcal{V}, \mathbf{A}, \mathbf{X}) $. The graph is comprised of a node set $ \mathcal{V} $ with a cardinality of $ n $, indexed as $ \{v_1, \dots, v_n\} $. The adjacency matrix $ \mathbf{A} \in \mathbb{R}^{n \times n} $ characterizes the structural relationships between nodes. We further define the degree matrix $ \mathbf{D} $ as a diagonal matrix, with $ \mathbf{D} = \text{diag}(d_1, \dots, d_n) $. Every node $ v_i $ is associated with a $ d $-dimensional feature vector $ \mathbf{x}_i \in \mathbb{R}^{d} $. When combined, these vectors form the feature matrix as $ \mathbf{X} = [\mathbf{x}_1, \dots, \mathbf{x}_n]^\top \in \mathbb{R}^{n \times d}$. The label of each node $y_i$ belongs to one among the $ C $ classes, enumerated as $ \{1,2,\dots, C\} $. We denote the one-hot encoding matrix of the labels as $\mathbf{B}$.

\paragraph{Semi-supervised node classification.}
In the context of our work, we tackle a semi-supervised node classification task. The entire node set $\mathcal{V} $ is divisibly partitioned into two discrete subsets: $ \mathcal{U} $, representing the unlabeled nodes, and $ \mathcal{L} $, encompassing the labeled nodes. Similarly, the original feature matrix $ \mathbf{X} $ is divided into $\mathbf{X}_{\mathcal{L}}$ and $\mathbf{X}_{\mathcal{U}}$, corresponding to the node sets they belong to. Our primary goal is to leverage the labeled subset $\mathcal{L}$ to predict class labels for the nodes in $\mathcal{U}$ with unknown labels. Beyond traditional classification tasks, the semi-supervised node classification task is confronted with a heightened challenge that there exists a predominant presence of unknown labels within the testing set compared to the limited known labels within the training set. This configuration echoes the settings explored in prior studies \cite{yang_revisiting_2016,shchur_pitfalls_2019,hu_open_2021}. 

\paragraph{Graph Neural Networks.}
Graph neural networks (GNNs) are a family of algorithms that extract structural information from graphs that encode graph-structured data into node representations or graph representations. They initialize each node feature representation with its attributes $\mathbf{H}^{(0)}=\mathbf{X}$ and then gradually update it by aggregating representations from its neighbors. Formally, given a graph $ \mathcal{G} = (\mathcal{V}, \mathbf{A}, \mathbf{X}) $, the $l$-th layer GNN is defined as:
\begin{align}
\label{eq:agg_upd}
\mathbf{H}^{(l)} \coloneqq \text{UPD} (\mathbf{H}^{(l-1)}, \text{AGG} (\mathbf{H}^{(l-1)},\mathbf{A}))),
\end{align}
where $\text{AGG}(\cdot)$ and $\text{UPD}(\cdot)$ denote the neighborhood aggregation function and the updating function respectively~\cite{gilmer_neural_2017}. 
As a result, the final layer output of a GNN, represented as $\mathbf{Z} = \mathbf{H}^{(l)} \in \mathbb{R}^{n \times C}$, serves as the predicted logits for different classes. To derive a prediction, one can select the class label associated with the highest logit for each node: 
\begin{align}
    \hat{y}_i= \operatorname*{arg\,max}_{c} {\mathbf{Z}_{i,c}}.
\end{align}



\paragraph{Decoupled GNNs.}
Prominent GNNs such as GCN typically incorporate learnable MLPs within the $\text{UPD}(\cdot)$ function across each layer. Nonetheless, recent studies~\cite{zheng_distribution_2022} suggest that decoupling message passing from feature transformation can deliver competitive performance when benchmarked against traditional GNNs. These so-called Decoupled GNNs (DeGNNs) are often characterized by their computational efficiency and are better to scale with larger graphs. Broadly, DeGNNs fall into two categories, distinguished by the sequence in which they conduct message passing and feature transformation. 

For example, the SGC model~\cite{wu_simplifying_2019} first undertakes multiple rounds of message passing before culminating with a trainable linear layer for prediction. It performs the message-passing step similar to GCN but without layer-wise linear transformation as:
\begin{align}
\mathbf{H}^{(l)} = {\left( \tilde{\mathbf{A}} \right)}^L \mathbf{X}, \mathbf{Z}=\mathbf{H}^{(l)}\mathbf{W}\label{eq:sgc},
\end{align}
where $\tilde{\mathbf{A}} = \mathbf{\hat{D}}^{-1/2} \mathbf{\hat{A}} \mathbf{\hat{D}}^{-1/2}$ and $\mathbf{\hat{A}} = \mathbf{A} + \mathbf{I}$ is the adjacency matrix with added self-loops. $\mathbf{W} \in \mathbb{R}^{d \times C}$ is the learnable weight matrix.

Conversely, the C\&S approach~\cite{huang_combining_2020} initiates training with an MLP exclusively on node attributes, devoid of the graph structure. It then propagates the resulting logits through the graph, refining prediction based on the graph structure as:
\begin{align}
\mathbf{\hat{Z}} = \text{MLP}(\mathbf{X}), \mathbf{Z}= \text{C\&S}(\mathbf{\hat{Z}},\mathbf{A})\label{eq:cs},
\end{align}
where $\mathbf{\hat{Z}}$ is the logits for the node classification tasks. For those keen on the specific formulation of $\text{C\&S}(\mathbf{Z_0},\mathbf{A})$, we have detailed it in the appendix.

\paragraph{Objective and Training Process}
To train GNNs, one must optimize their weight matrices to fit on the dataset. Surrogate measures such as cross entropy, depicted in \Eqref{eq:loss}, are employed to iteratively adjust the model weights to reduce discrepancies between its predictions and the true labels on the labeled subset $\mathcal{L}$ of the graph.
\begin{align}
    \text{Loss} = -\frac{1}{|\mathcal{L}|} \sum_{v_i \in \mathcal{L}} \log \left( \frac{e^{\mathbf{Z}_{i,y_i}}}{\sum_{j=1}^{C} e^{\mathbf{Z}_{i,j}}} \right). \label{eq:loss}
\end{align}

\section{Unpacking What GNNs Learn on Text-Attributed Graphs}
In this section, our goal is to gain insights into the inner workings of GNNs, especially during training for node classification on graphs that have text attributes. We will first discuss common text encoding methods used to convert textual information into a format suitable for machine learning. Then, we will examine the behavior of weight matrices in two GNN models: SGC and GCN. Insights from this exploration will pave the way for our proposed method, which does not require to optimize the loss function to fit the data.

\subsection{Quasi-orthogonal node attributes}
On a TAG, nodes contain text descriptions that highlight their unique characteristics. But to make this text data useful for machine learning algorithms, we convert it into vectors. Node attributes in such a graph are commonly encoded using methods like Bag-of-Words (BOW) and TF-IDF~\cite{yang_revisiting_2016,shchur_pitfalls_2019,hu_open_2021}. The essence of these methods is to construct a vocabulary from word tokens and then encode documents based on word token occurrences. For expansive vocabularies, such encoding tends to be sparse. This sparsity implies that two documents with differing lexicons are likely to yield encodings that are orthogonal. We term this attribute behavior as \textit{quasi-orthogonality} (QO).
\begin{figure}[h!]
    \centering
    \includegraphics[width=1.0\linewidth]{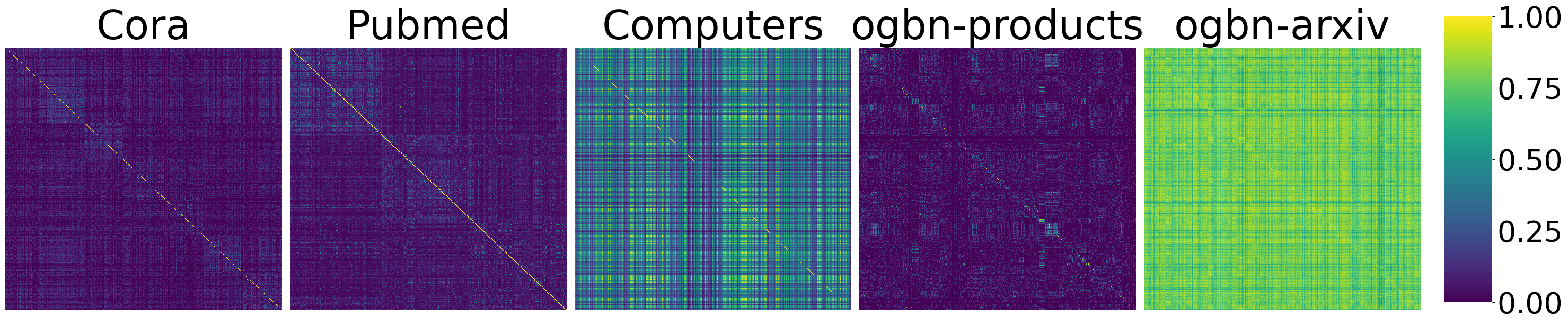}
    \caption{Heatmap of the inner product of node attributes on TAG.\vspace{-1mm}}
    \label{fig:heatmap}\vspace{-1mm}
\vspace{-1mm}
\end{figure}
We sought to empirically ascertain the QO of node attributes across various TAG. Our analysis spanned five distinct datasets: Cora (BOW encoded), Pubmed (TF-IDF encoded)~\cite{yang_revisiting_2016}, Computers (BOW encoded but with relatively smaller vocabulary)~\cite{shchur_pitfalls_2019}, OGBN-Products (BOW encoded followed by principle component analysis), and OGBN-Arxiv (encoded via word2vec averaging)~\cite{hu_open_2021}.

The heatmap in ~\Figref{fig:heatmap} plots the inner products of node attributes. Both the x and y axes denote node indices, whereas the color gradient signifies the magnitude of their inner product. We further order nodes by their true labels, $y$, ensuring nodes from identical classes cluster adjacently. 

In the datasets of Cora, Pubmed, and OGN-Product, the heatmap clearly exhibits QO. The inner product between attributes of any node pair is almost negligible. Moreover, brighter blocks along the diagonal indicate that nodes within the same classes have a tendency to exhibit higher inner products compared to other nodes. In contrast, the OGBN-Arxiv dataset doesn't demonstrate this QO trait. This deviation can be attributed to the use of average word embeddings for node attribute generation, which can compromise the QO among nodes. Interestingly, despite being encoded with BOW, the Computers dataset lacks prominent QO node attributes. This might stem from its relatively smaller vocabulary size. In the experiment section, we will illustrate that the QO property plays a critical role in determining the efficacy of our proposed trainless methods compared to the trained approaches.

\subsection{What SGC learns}
In this section, we probe deeper into the learning dynamics of the SGC model when trained via gradient descent. Notably, the SGC model comprises a sole trainable weight matrix, $\mathbf{W} = (\mathbf{W}_{:,1},\dots,\mathbf{W}_{:,C}) \in \mathbb{R}^{d \times C}$, as illustrated in \Eqref{eq:sgc}. Our primary focus lies in understanding the interplay between the node attributes from the training set and this weight matrix.

The logit $\mathbf{Z}_{i,c}$ for node $i$ concerning class $c$ is computed by the inner product $\mathbf{Z}_{i,c} = \mathbf{x}_i \cdot \mathbf{W}_{:,c}$. For a prediction to be accurate, we would anticipate that the inner product with the weight vector corresponding to the true class surpasses those of other classes. Given the QO observed in node attributes, this phenomenon might be even more pronounced.
\begin{figure*}[h!]
    \centering
    \includegraphics[width=1.0\linewidth]{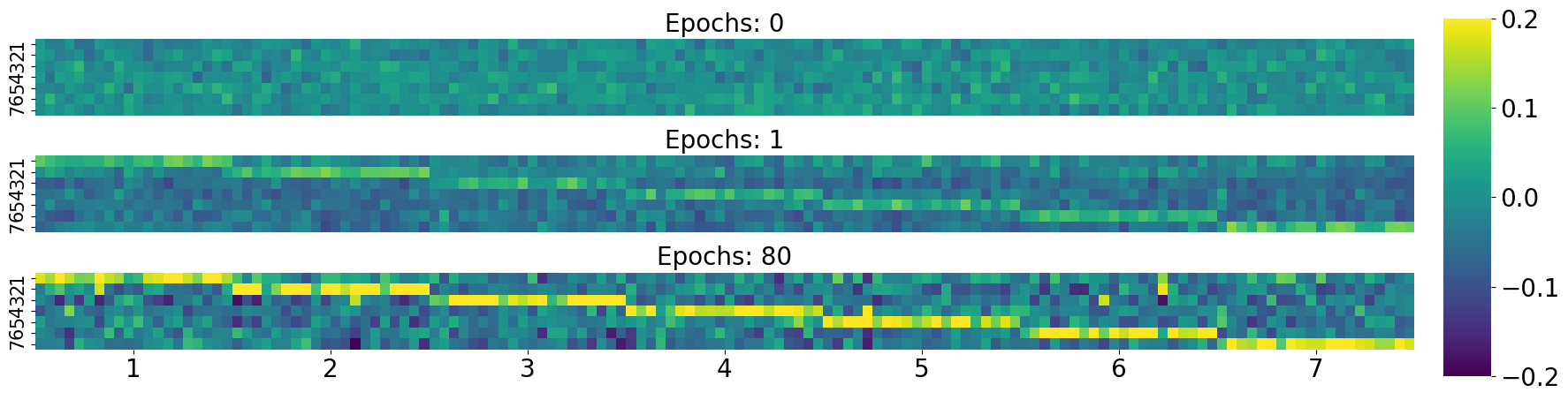}
    \caption{Heatmap depicting the evolution of inner products between node attributes and the weight vectors across various training epochs on the Cora dataset.\vspace{-2mm}}
    \label{fig:sgc_training}\vspace{-2mm}
\end{figure*}
We base our experiment on the Cora dataset, characterized by $7$ label classes with $20$ labeled nodes for each class~\cite{yang_revisiting_2016}. We evaluate the inner product of each column vector $\mathbf{W}_{:,c}$ of the weight matrix (for $1\leq c \leq 7$) against the node attributes $\mathbf{x}_i$ for nodes $v_i \in \mathcal{L}$. This results in a heatmap of $7$ rows and $20 \times 7 = 140$ columns, with nodes of identical labels grouped together. The progression of this heatmap across various epochs during training is displayed in \Figref{fig:sgc_training}.

The evolving heatmaps reveal a pattern: as training progresses, the weight matrix inclines to heighten the inner product between a node $v_i$'s attribute and its corresponding class's weight vector, $\mathbf{W}_{:,y_i}$, while diminishing the product with other classes. This amplifies the logit for the true class $y_i$, suppressing logits for other classes towards zero, subsequently reducing the loss in \Eqref{eq:loss}. This observation serves as a foundation for our subsequent proposal, where we seek to derive the weight matrix directly from the aggregation of node attributes.


\subsection{What GCN learns}
\vspace{-3mm}
\begin{table}[h]
\centering
\caption{Comparison of accuracy between SGC, $2$-layer GCN, and $2$-layer GCN with the second layer frozen.}
\label{tab:gcn}
\begin{tabular}{l|ccc}
\toprule
Dataset & Cora & Citeeer & Pubmed \\
\midrule
SGC & $81.00$ & $71.90$ & $78.90$\\ \midrule
$2$-layer GCN & $81.50$ & $71.40$ & $78.50$\\ 
second-layer-frozen GCN & $80.40$ & $70.50$ & $77.80$\\
\bottomrule
\end{tabular}
\vspace{-3mm}
\end{table}
We previously observed the SGC's propensity to optimize its weight matrix, ensuring a heightened inner product between node attributes and the corresponding class's weight vector. This section delves into discerning whether GNNs, particularly those with non-linearities and multiple layers, exhibit analogous learning dynamics. We direct our focus to the popularly employed $2$-layer GCN.

We first rewrite the Equation \ref{eq:agg_upd} for GCN as:
\begin{align}
    \mathbf{H}^{(1)} = \sigma(\tilde{\mathbf{A}} \mathbf{X} \mathbf{W}^{(1)}), \quad
    \mathbf{H}^{(2)} = \tilde{\mathbf{A}} \mathbf{H}^{(1)} \mathbf{W}^{(2)}.
\end{align}

Interestingly, the one-layer model, such as the SGC, achieves comparable performance with the $2$-layer GCN, as shown in \autoref{tab:gcn}. This leads us to question the necessity of simultaneously training weight matrices for both layers. To explore this, we train a $2$-layer GCN, but keep the second layer's weight matrix frozen, preserving its initial state. The outcomes, shown in \autoref{tab:gcn}, reveal that performance remains robust even with a static second layer. This suggests that in scenarios where node attributes are sufficiently learned by a linear model like SGC, the MLP integrated within GCNs might be redundant.

Pursuing this inquiry, with the second layer's weight matrix still frozen, we undertake an experiment analogous to our earlier one on SGC. Unlike our previous focus on the inner product between node attributes $\mathbf{X}$ and weight vectors $\mathbf{W}_{:,c}$, we now assess the correlation between the node representation $\mathbf{H}^{(1)}$ from the GCN's first layer and the second layer's weight matrix $\mathbf{W}^{(2)}$. Specifically, we calculate the inner product between each training node's representation $\mathbf{H}^{(1)}_{i,:}$ and the column vector $\mathbf{W}^{(2)}_{:,c}$ of the weight matrix, then similarly represent this using a heatmap in \Figref{fig:gcn_training} (in Appendix).

Interestingly, the first layer of the GCN appears to forge a relationship with the second layer's weight matrix that mirrors SGC's dynamics. Notably, the inner product between the node representation $\mathbf{H}^{(1)}_{i,:}$ and its associated class's weight vector $\mathbf{W}^{(2)}_{:,y_i}$ is markedly higher than with the weight vectors of other classes. Throughout the training phase, the GCN's first layer essentially learns to project node attributes from different classes into different subspaces. These subspaces are inherently defined by the corresponding randomly initialized weight vector $\mathbf{W}^{(2)}_{:,c}$ of the second layer. Given that randomly initialized vectors tend to be QO with high probability \cite{dong_pure_2023}, this fosters a high inner product between $\mathbf{H}^{(1)}_{i,:}$ and $\mathbf{W}^{(2)}_{:,y_i}$, but nearly nullifies the product with other classes. This matches with our SGC observations.

\section{Method}
Previous analyses of SGC and GCN highlight that in cases where node attributes from different classes are nearly orthogonal, gradient descent training tends to align the weight vectors with corresponding class node attributes. It's also observed that for GCNs, freezing the last layer suggests that a sole linear layer can suffice for TAG. Based on these insights, we introduce a simple yet efficient method, referred to as \our. This method creates the linear weight matrix directly from node attributes, eliminating the need for the usual gradient descent training in GNNs, while still being suitable for inference. It's applicable to linear classification models and any De-GNN with a linear layer. Further details about the implementation can be found in Appendix.

\subsection{Building the Weight Matrix}
\begin{figure}[h]
    \centering
    \includegraphics[width=1.0\linewidth]{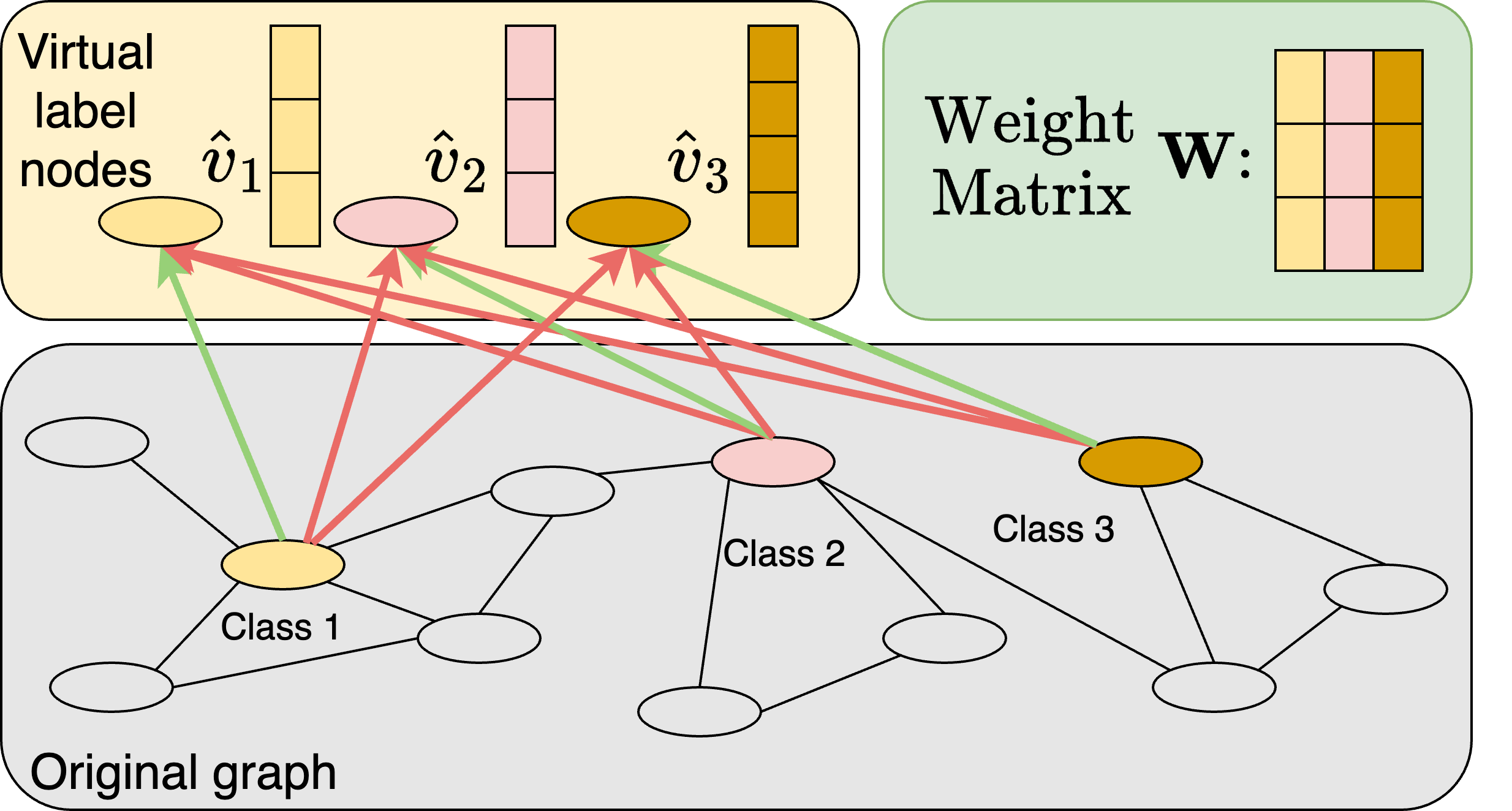}\vspace{-2mm}\caption{This figure outlines the process for obtaining the weight matrix $ \mathbf{W} $ in \our. Initially, virtual label nodes are added for each class label. These nodes are then connected to labeled nodes sharing the same class, depicted by \textcolor{MyGreen}{green lines}. Additionally, virtual label nodes are connected to all other labeled nodes, represented by \textcolor{MyRed}{red lines}, with an assigned edge weight $ \omega $. A single round of message passing updates the representation of the virtual label nodes, providing the desired weight matrix $ \mathbf{W} $.\vspace{-1mm}}\label{fig:tgnn}\vspace{-3mm}
\end{figure}

\paragraph{Virtual label nodes.}To construct a weight matrix $ \mathbf{W} \in \mathbb{R}^{d \times C} $ applicable to any linear model, it's essential to formulate the weight vectors $ \mathbf{W}_{:,c} \in \mathbb{R}^{d} $ for $ 1 \leq c \leq C $. It begins by adding $ C $ \textit{virtual label nodes} into the original graph as $ \mathcal{V}_{new} = \mathcal{V} \cup \{\hat{v}_{c}| 1 \leq c \leq C\} $. Every virtual label node $ \hat{v}_{c} $ symbolizes the corresponding class $ c $, initialized as a zero vector $ \mathbf{0} \in \mathbb{R}^{d} $. Subsequent to this, each virtual label node is connected to labeled nodes $ \mathcal{L} $ possessing the matching labels. For example, the virtual label node $ \hat{v}_{c} $ connects with all nodes from the training set labeled as $ c $, $ \{ v_i \in \mathcal{L}|y_i=c\} $. The \textcolor{MyGreen}{green lines} in Figure \ref{fig:tgnn} depict such connections.

\paragraph{Message passing.}In this newly formed graph with virtual label nodes, the weight matrix is constructed by executing a single round of message passing, which essentially updates the representation of the virtual label nodes. The updated virtual label nodes representation then defines $ \mathbf{W} $:
\begin{align}
\label{eq:mp}
    \mathbf{W}^\top = \mathbf{B}_{\mathcal{L}}^\top \mathbf{X}_{\mathcal{L}},
\end{align}
where $ \mathbf{X}_{\mathcal{L}} $ represents the node attributes from the labeled node sets $ \mathcal{L} $. The $ \mathbf{B}_{\mathcal{L}} \in \mathbb{R}^{|\mathcal{L}| \times C} $ is a one-hot encoding matrix of the labels of $ \mathcal{L} $, acting as the incidence matrix between virtual label nodes $ \{\hat{v}_{c}| 1 \leq c \leq C\} $ and labeled nodes $ \mathcal{L} $. Through this approach, the weight vectors are essentially constructed based on the node attributes from the corresponding classes, with the aim to maximize the inner product between the node attribute and the weight vector of the same class.

\paragraph{Connecting nodes with different labels.} Adjusting the weight vectors based on the node attributes from the same classes maximizes the inner product but fails to minimize the inner product for nodes from different classes. To address this, we extend the connections of virtual label nodes to other labeled nodes in the training set (\textcolor{MyRed}{red lines} in Figure \ref{fig:tgnn}). Contrarily to the connections within the same class nodes, we assign an edge weight $ \omega \in \mathbb{R} $ as a hyperparameter to the newly formed connections between virtual label nodes and differently labeled nodes. More formally, the weight matrix is computed as:
\begin{align}
\label{eq:neg_mp}
    \mathbf{W}^\top = (\mathbf{B}_{\mathcal{L}}-\frac{\omega}{C}\mathbf{1})^\top \mathbf{X}_{\mathcal{L}}.
\end{align}
In this equation, each entry of the one-hot encoding matrix $ \mathbf{B}_{\mathcal{L}} $ is subtracted by $ \frac{\omega}{C} $. By setting $ \omega $ to a negative value, we can achieve a weight matrix that not only maximizes the inner product of the weight vectors and node attributes from the same class but also minimizes the inner product from different classes.

\paragraph{Inference.}
After obtaining the trainless weight matrix $ \mathbf{W} $, we proceed to an accurate and efficient computation of the logits for the unlabeled node sets $ \mathcal{U} $. The logits are calculated with various backbone models as detailed below:
\begin{itemize}
    \item \textbf{Trainless Linear:} The logits $ \mathbf{Z} $ are directly computed using the expression:
    \begin{equation}
        \mathbf{Z} = \mathbf{X} \mathbf{W},
    \end{equation}
    where $ \mathbf{X} $ is the original feature matrix.

    \item \textbf{Trainless SGC:} The logits $ \mathbf{Z} $ are inferred using the updated node representations $ \mathbf{H}^{(l)} $ with:
    \begin{equation}
        \mathbf{Z} = \mathbf{H}^{(l)} \mathbf{W},
    \end{equation}

    \item \textbf{Trainless C\&S:} The logits $ \mathbf{Z} $ are obtained using the C\&S function applied to the product of the original feature matrix $ \mathbf{X} $ and the weight matrix $ \mathbf{W} $ alongside the adjacency matrix $ \mathbf{A} $:
    \begin{equation}
        \mathbf{Z}= \text{C\&S}(\mathbf{X} \mathbf{W},\mathbf{A}).
    \end{equation}
\end{itemize}

\subsection{A View from Linear Regressions}


In this section, we explore the equivalence between our approach and a linear classifier trained through gradient descent with a cross-entropy loss. We provide a rationale for our method, viewing it through the lens of linear regression. The discussion begins by outlining the following assumptions related to the semi-supervised node classification task:
\vspace{-1mm}
\begin{assumption}
\label{assumption}
Consider a graph $ \mathcal{G} = (\mathcal{V}, \mathbf{A}, \mathbf{X})$, where $\mathbf{X} \in \mathbb{R}^{n \times d}$. We propose that:
\begin{enumerate}
    \item The model is over-parametrized as the number of features $d$ is adequately large, i.e., $d > |\mathcal{L}|$, where $\mathcal{L}$ is the training/labeled set.
    \item The row vectors of the feature matrix are orthogonal as $\mathbf{X}_\mathcal{L}\mathbf{X}_\mathcal{L}^\top=\mathbf{I}$.
\end{enumerate}
\end{assumption}


These assumptions are modest and align with most real-world scenarios. The first assumption pertinently applies to the majority of TAG created using shallow text encoding methods like BOW or TF-IDF. Here, the number of features is contingent on the size of the vocabulary. Additionally, in many semi-supervised node classification task setups, the number of nodes in the training set is often comparatively low \cite{yang_revisiting_2016,shchur_pitfalls_2019,hu_open_2021}, further accentuating the model's over-parameterization (refer to Table \ref{tab:stats}). The second assumption, grounded in previous observations, suggests that node attributes are likely orthogonal to each other (see \Figref{fig:heatmap}).

In a simplified form, the classification problem in \Eqref{eq:loss} can be naively converted into a linear regression problem with the least squared loss:
\begin{align*}
    \mathbf{\hat{W}} &= \arg \min_\mathbf{W} \| \mathbf{X}_{\mathcal{L}} \mathbf{W} - \mathbf{B}_{\mathcal{L}} \|_2\label{eq:regression}.
\end{align*}
Thanks to the over-parameterization, the training loss can be reduced to zero by $\mathbf{X}_{\mathcal{L}} \mathbf{\hat{W}} = \mathbf{B}_{\mathcal{L}}$.
Within this framework, our method is viewed as an effort to transpose the $\mathbf{X}_{\mathcal{L}}$ term to the right-hand side. However, while our assumption holds that $\mathbf{X}_\mathcal{L}\mathbf{X}_\mathcal{L}^\top=\mathbf{I}$, it does not assert that $\mathbf{X}_\mathcal{L}^\top \mathbf{X}_\mathcal{L}=\mathbf{I}$. As a result, our solution cannot be straightforwardly derived from this linear regression format.

To robustly affirm the effectiveness of our methodology, we employ the minimum-norm interpolation method. Given the aforementioned assumptions, we can reformulate the weight matrix $\mathbf{W}$ derived by our method in \Eqref{eq:mp} as:
\begin{align}
    \mathbf{W} = \mathbf{X}_{\mathcal{L}}^\top \mathbf{B}_{\mathcal{L}} = \mathbf{X}_{\mathcal{L}}^\top \mathbf{I} \mathbf{B}_{\mathcal{L}} = \mathbf{X}_{\mathcal{L}}^\top (\mathbf{X}_{\mathcal{L}} \mathbf{X}_{\mathcal{L}}^\top)^{-1} \mathbf{B}_{\mathcal{L}}.
\end{align}
Essentially, the formulation above acts as an estimator for the linear regression task, addressed by the minimum-norm interpolation method \cite{wang_benign_2023}:
\begin{align}
\label{eq:linear_loss}
\mathbf{\hat{W}} &= \arg \min_\mathbf{W} \| \mathbf{W} \|_2 \ , \  \text{s.t.} \  \mathbf{X}_{\mathcal{L}} \mathbf{W} = \mathbf{B}_{\mathcal{L}}.
\end{align}
Moreover, the weight matrix $\mathbf{W}$ acquired through the minimum-norm interpolation and in \Eqref{eq:mp} is equivalent to that achieved by other standard training methods, including SVMs or gradient descent with diverse losses such as cross-entropy, with sufficient over-parameterization \cite{wang_benign_2023}. 
This equivalency bolsters the legitimacy of our methodology.

In summary, our technique essentially transforms a convex optimization problem lacking closed-form solutions (\Eqref{eq:loss}) into a linear regression (\Eqref{eq:linear_loss}) with a closed-form solution (\Eqref{eq:mp}). This transformation is grounded on the distinctive sparse encodings of the TAG data.

%% file: body/experiment.tex
\section{Experiments}

\begin{table*}[ht]
    \centering
    \caption{Results of semi-supervised node classification on benchmark datasets, evaluated by accuracy. The format is average score ± standard deviation. The top three models are colored by \textbf{\textcolor{red}{First}}, \textbf{\textcolor{blue}{Second}}, \textbf{\textcolor{violet}{Third}}.}
    \resizebox{\textwidth}{!}{%
    \begin{tabular}{lccccccccc}
    \toprule
        & \textbf{Cora} & \textbf{Citeseer} & \textbf{Pubmed} & \textbf{CS} & \textbf{Physics} & \textbf{Computers} & \textbf{Photo} & \textbf{OGBN-Products} & \textbf{OGBN-Arxiv}\\\midrule
    \textbf{LP} & $68.00 {\scriptstyle \pm 0.00}$ & $45.30 {\scriptstyle \pm 0.00}$ & $63.00 {\scriptstyle \pm 0.00}$ & $73.60 {\scriptstyle \pm 3.90}$ & $86.60 {\scriptstyle \pm 2.00}$ & $70.80 {\scriptstyle \pm 8.10}$ & $72.60 {\scriptstyle \pm 11.10}$ & $74.34 {\scriptstyle \pm 0.00}$ & $68.32 {\scriptstyle \pm 0.00}$ \\
    \textbf{GCN} & $80.94 {\scriptstyle \pm 0.45}$ & $69.24 {\scriptstyle \pm 0.74}$ & $76.62 {\scriptstyle \pm 0.30}$ & $90.01 {\scriptstyle \pm 1.08}$ & $92.10 {\scriptstyle \pm 1.42}$ & \firstt{82.46}{1.66} & \thirdt{88.10}{1.48} & $75.64 {\scriptstyle \pm 0.21}$ & \firstt{71.74}{0.29} \\
    \textbf{SAGE} & $80.03 {\scriptstyle \pm 0.70}$ & $69.27 {\scriptstyle \pm 0.99}$ & $76.59 {\scriptstyle \pm 0.32}$ & $89.76 {\scriptstyle \pm 0.61}$ & $91.18 {\scriptstyle \pm 1.52}$ & $81.19 {\scriptstyle \pm 2.03}$ & $87.58 {\scriptstyle \pm 2.21}$ & \secondt{78.29}{0.16} & \thirdt{71.49}{0.27} \\ \midrule
    \textbf{Linear} & $59.20 {\scriptstyle \pm 0.20}$ & $60.70 {\scriptstyle \pm 0.10}$ & $72.70 {\scriptstyle \pm 0.16}$ & $87.64 {\scriptstyle \pm 0.68}$ & $87.83 {\scriptstyle \pm 1.16}$ & $57.55 {\scriptstyle \pm 5.23}$ & $76.51 {\scriptstyle \pm 2.59}$ & $46.45 {\scriptstyle \pm 0.52}$ & $41.78 {\scriptstyle \pm 0.23}$ \\
    \textbf{SGC} & \thirdt{81.00}{0.00} & $71.90 {\scriptstyle \pm 0.10}$ & $78.90 {\scriptstyle \pm 0.00}$ & $90.60 {\scriptstyle \pm 0.96}$ & $92.66 {\scriptstyle \pm 0.89}$ & \secondt{82.33}{1.39} & \firstt{89.64}{2.05} & $70.67 {\scriptstyle \pm 0.20}$ & $67.63 {\scriptstyle \pm 0.32}$ \\
    \textbf{C\&S} & $78.40 {\scriptstyle \pm 0.00}$ & $69.70 {\scriptstyle \pm 0.00}$ & $75.40 {\scriptstyle \pm 0.00}$ & \secondt{91.32}{1.29} & $92.13 {\scriptstyle \pm 2.57}$ & $70.70 {\scriptstyle \pm 11.01}$ & $85.09 {\scriptstyle \pm 4.02}$ & \firstt{82.54}{0.03} & $71.26 {\scriptstyle \pm 0.01}$ \\ \midrule
    \textbf{Trainless Linear} & $59.10 {\scriptstyle \pm 0.00}$ & $63.10 {\scriptstyle \pm 0.00}$ & $72.40 {\scriptstyle \pm 0.00}$ & $87.97 {\scriptstyle \pm 0.66}$ & $88.06 {\scriptstyle \pm 1.05}$ & $62.12 {\scriptstyle \pm 1.84}$ & $73.38 {\scriptstyle \pm 2.60}$ & $37.12 {\scriptstyle \pm 0.00}$ & $41.57 {\scriptstyle \pm 0.00}$ \\
    \textbf{Trainless SGC} & $79.60 {\scriptstyle \pm 0.00}$ & \thirdt{73.00}{0.00} & $76.40 {\scriptstyle \pm 0.00}$ & \thirdt{91.22}{0.56} & $92.74 {\scriptstyle \pm 1.37}$ & $77.32 {\scriptstyle \pm 1.73}$ & $83.45 {\scriptstyle \pm 1.73}$ & $60.48 {\scriptstyle \pm 0.00}$ & $61.71 {\scriptstyle \pm 0.00}$ \\
    \textbf{Trainless C\&S} & $77.90 {\scriptstyle \pm 0.00}$ & $68.40 {\scriptstyle \pm 0.00}$ & $75.30 {\scriptstyle \pm 0.00}$ & $88.89 {\scriptstyle \pm 0.54}$ & \secondt{93.12}{0.76} & $78.91 {\scriptstyle \pm 1.41}$ & $87.06 {\scriptstyle \pm 2.32}$ & $77.27 {\scriptstyle \pm 0.00}$ & $69.52 {\scriptstyle \pm 0.00}$ \\ \midrule
    \multicolumn{10}{l}{\textit{Use both training and validation labels}} \\
    \textbf{Trainless Linear } & $68.20 {\scriptstyle \pm 0.00}$ & $71.20 {\scriptstyle \pm 0.00}$ & \thirdt{79.20}{0.00} & $88.99 {\scriptstyle \pm 0.59}$ & $89.33 {\scriptstyle \pm 0.47}$ & $68.28 {\scriptstyle \pm 1.21}$ & $76.17 {\scriptstyle \pm 1.80}$ & $37.57 {\scriptstyle \pm 0.00}$ & $42.84 {\scriptstyle \pm 0.00}$ \\
    \textbf{Trainless SGC } & \secondt{82.70}{0.00} & \firstt{77.20}{0.00} & \firstt{81.30}{0.00} & \firstt{91.38}{0.67} & \thirdt{92.93}{0.61} & $79.21 {\scriptstyle \pm 0.50}$ & $84.75 {\scriptstyle \pm 2.54}$ & $60.51 {\scriptstyle \pm 0.00}$ & $62.56 {\scriptstyle \pm 0.00}$ \\
    \textbf{Trainless C\&S } & \firstt{83.80}{0.00} & \secondt{73.20}{0.00} & \secondt{79.90}{0.00} & $88.16 {\scriptstyle \pm 0.40}$ & \firstt{93.49}{0.20} & \thirdt{81.67}{0.96} & \secondt{88.69}{0.78} & \thirdt{77.90}{0.00} & \secondt{71.70}{0.00} \\
    \bottomrule
    \end{tabular}
}
\label{tab:main}
\end{table*}

In this section, we evaluate \our on different benchmark datasets. We start by evaluating our method on nine TAG datasets with different scales.

\subsection{Experimental setups}
\paragraph{Datasets.}We select nine commonly used TAGs as our benchmarks. We use the citation network Planetoid datasets \cite{yang_revisiting_2016}, including Cora, Citeseer and Pubmed. We also use the datasets introduced by \cite{shchur_pitfalls_2019}, including two Coauthor datasets, CS and Physics, and the Amazon co-purchase networks, Computers and Photo. We further include two OGB datasets \cite{hu_open_2021} such as OGBN-Products and OGBN-Arxiv. 
\vspace{-2mm}
\paragraph{Baseline models.}We select LP \cite{zhu_semi-supervised_2003} as the graph-Laplacian baseline. We choose GCN \cite{kipf_semi-supervised_2017} and SAGE \cite{hamilton_inductive_2018}, two of the most representative GNNs, as the non-linear baseline models. We then select logistic regression (denoted as \textbf{Linear}), \textbf{SGC}, and \textbf{C\&S} as the linear baseline models. We implement three types of \our, including \textbf{Trainless Linear}, \textbf{Trainless SGC}, and \textbf{Trainless C\&S}, corresponding to the trainless versions of the linear baseline models.
\vspace{-2mm}
\paragraph{Evaluation protocols.}We evaluate the models based on the accuracy of the test set. For datasets with predefined train/test splits (Planetoid and OGBN datasets), we follow their splits and run the evaluation $10$ times for different model initializations. For datasets without predefined splits, we follow previous studies of semi-supervised node classification tasks \cite{shchur_pitfalls_2019}, splitting the labeled nodes into training/validations/testing sets for $10$ splits. For each split, we randomly pick $20$/$30$ nodes from each class label as the training/validation sets and leave the rest as the testing sets. We then evaluate the model performance on $10$ splits and report the average and standard deviations of the accuracy. 
\vspace{-2mm}
\begin{figure*}[h]
     \centering
     \begin{subfigure}[b]{0.3\textwidth}
         \centering
         \includegraphics[width=\textwidth]{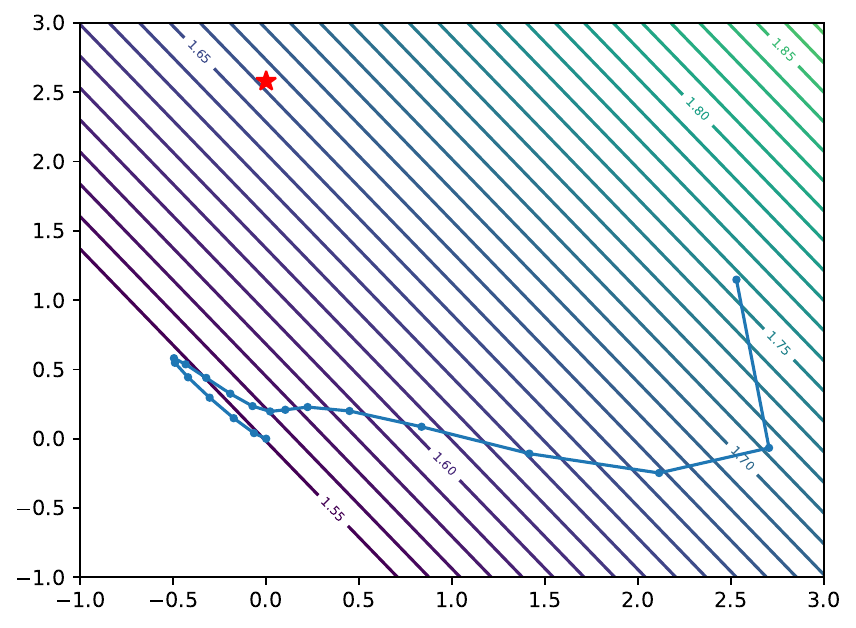}
         \caption{Training loss.}
         \label{fig:train_loss}
     \end{subfigure}
     \hfill
     \begin{subfigure}[b]{0.3\textwidth}
         \centering
         \includegraphics[width=\textwidth]{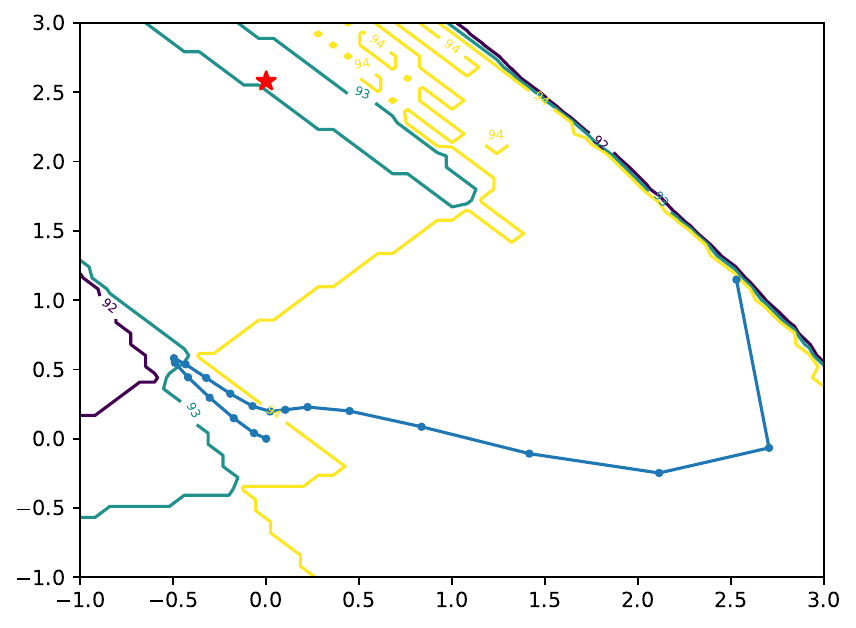}
         \caption{Training accuracy.}
         \label{fig:train_acc}
     \end{subfigure}
     \hfill
     \begin{subfigure}[b]{0.3\textwidth}
         \centering
         \includegraphics[width=\textwidth]{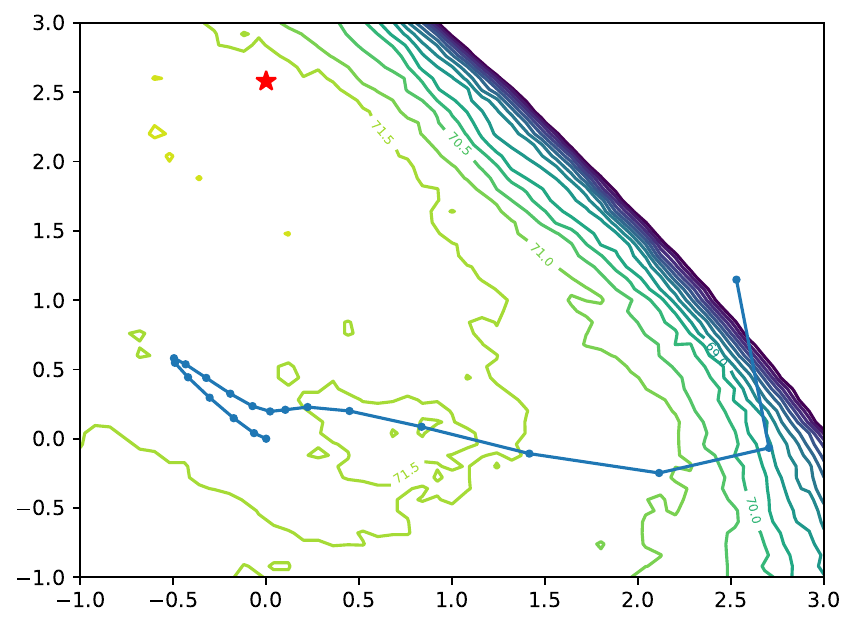}
         \caption{Testing accuracy.}
         \label{fig:test_acc}
     \end{subfigure}
        \caption{The loss/accuracy landscape while training SGC on Citeseer. The red star (\textcolor{red}{$\star$}) denotes Trainless SGC.}
        \label{fig:converge}
\end{figure*}
\subsection{Results}
We present our results in two parts. Initially, we fit \our using only the labeled nodes from the training set, which is the conventional approach for training GNNs. This is favorable to baseline models. Subsequently, we include labeled nodes from both training and validation sets to fit the model. This comparison remains fair as \our, with only tunable hyperparameter $\omega$, is less likely to overfit on training data, eliminating the need for an exclusive validation set to prevent overfitting. Conversely, typical neural networks, especially in over parameterized domains, are prone to overfitting to zero loss, necessitating a validation set for model generalization. The results of both scenarios are shown in Table \ref{tab:main}.
\vspace{-2mm}
\paragraph{\our on training sets.}
When fit on the training set, \our achieves comparable performance across various benchmarks to trained models. Specifically, on Cora and Pubmed, \our matches the performance of trained models, and notably surpasses them on Citeseer, CS, and Physics by 0.3\% to 2.6\%. However, on four other datasets, our trainless method trails slightly. Among these, Computers and Photo exhibit a weak quasi-orthogonal property, while the OGB datasets have more training labels relative to node attribute dimensions, affirming the importance of quasi-orthogonal property and over-parameterization in Assumption \ref{assumption} for our method.
\vspace{-2mm}
\paragraph{\our on both training and validation sets.}
The inclusion of validation labels is a distinct advantage of \our, further enhancing its performance. Specifically, on the three Planetoid and two Coauthor datasets, \our outperforms all baseline models significantly when fitted with both training and validation labels. Remarkably, our trainless models even exceed the performance of trained GCN/SAGE models, which possess higher expressiveness with non-linear MLPs. For the remaining datasets, including the validation set also boosts \our's performance, aligning it with that of trained models.
\subsection{Trained vs Trainless weight matrix}
We extend our analysis by contrasting the weight matrix obtained through our method with that learned via a standard gradient descent process under cross entropy. We illustrate the learning trajectory of \textbf{SGC} over the initial $20$ epochs alongside the fitted \textbf{Trainless SGC} on the Citeseer dataset. The loss and accuracy landscape is shown in Figure \ref{fig:converge}. In Figure \ref{fig:train_loss}, the training loss of SGC steadily diminishes through optimization towards a minimal point, a trend guaranteed by the convex nature of the loss function. Conversely, while Trainless SGC settles at a point with relatively higher loss, its accuracy on both training (Figure \ref{fig:train_acc}) and testing (Figure \ref{fig:test_acc}) sets achieves a level comparable to the trained model. This insight implies that attaining high accuracy does not indispensably hinge on the optimization of surrogate loss. A well-generalizable model can indeed be identified without resorting to gradient descent training.

\subsection{Varying attribute dimensions}
\vspace{-4mm}
\begin{figure}[h]
    \centering
    \resizebox{1.0\linewidth}{!}{%
    \includegraphics[width=1.0\linewidth]{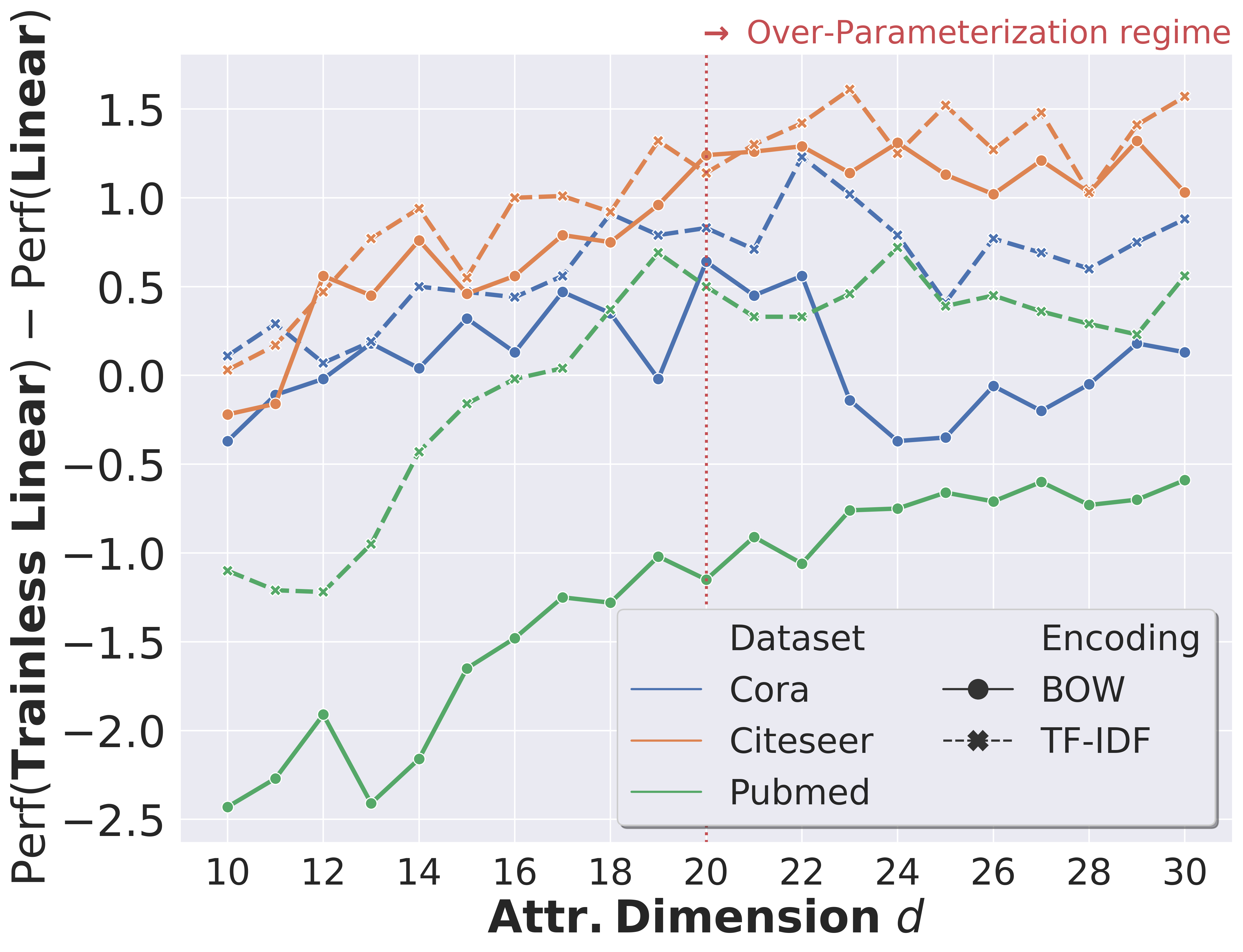}
    }
    \caption{
Performance comparison between \textbf{Trainless Linear} and \textbf{Linear} across varying attribute dimensions and textual encodings, with a consistent training set of $20$ labeled nodes. Attribute dimensions greater than $20$ (\textit{i.e.,} $d>20$) represent an over-parameterization regime.\vspace{-2mm}}
    \label{fig:parameterization}
    \vspace{-2mm}
\end{figure}
Recalling Assumption \ref{assumption}, we assume that a large attribute dimension is crucial for over-parameterization, enabling our closed-form solution to approximate the optimal point effectively. We test this by varying the attribute dimensions of node attributes in the three Planetoid datasets~\cite{chen_exploring_2023}, using both BOW and TF-IDF text encodings. We compare the performance of a \textbf{Trainless Linear} and a trained logistic regression (\textbf{Linear}) across these encodings. The results, presented in Figure \ref{fig:parameterization}, indicate that increasing attribute dimensions enhances the performance of the \textbf{Trainless Linear} model over the trained one. This supports the effectiveness of our trainless approach in semi-supervised node classification with sparse labels and lengthy text-encoded node attributes.
\vspace{-3mm}
\subsection{Beyond homophilous graphs}
\begin{figure}[h]
    \centering
    \resizebox{1.0\linewidth}{!}{%
    \includegraphics[width=1.0\linewidth]{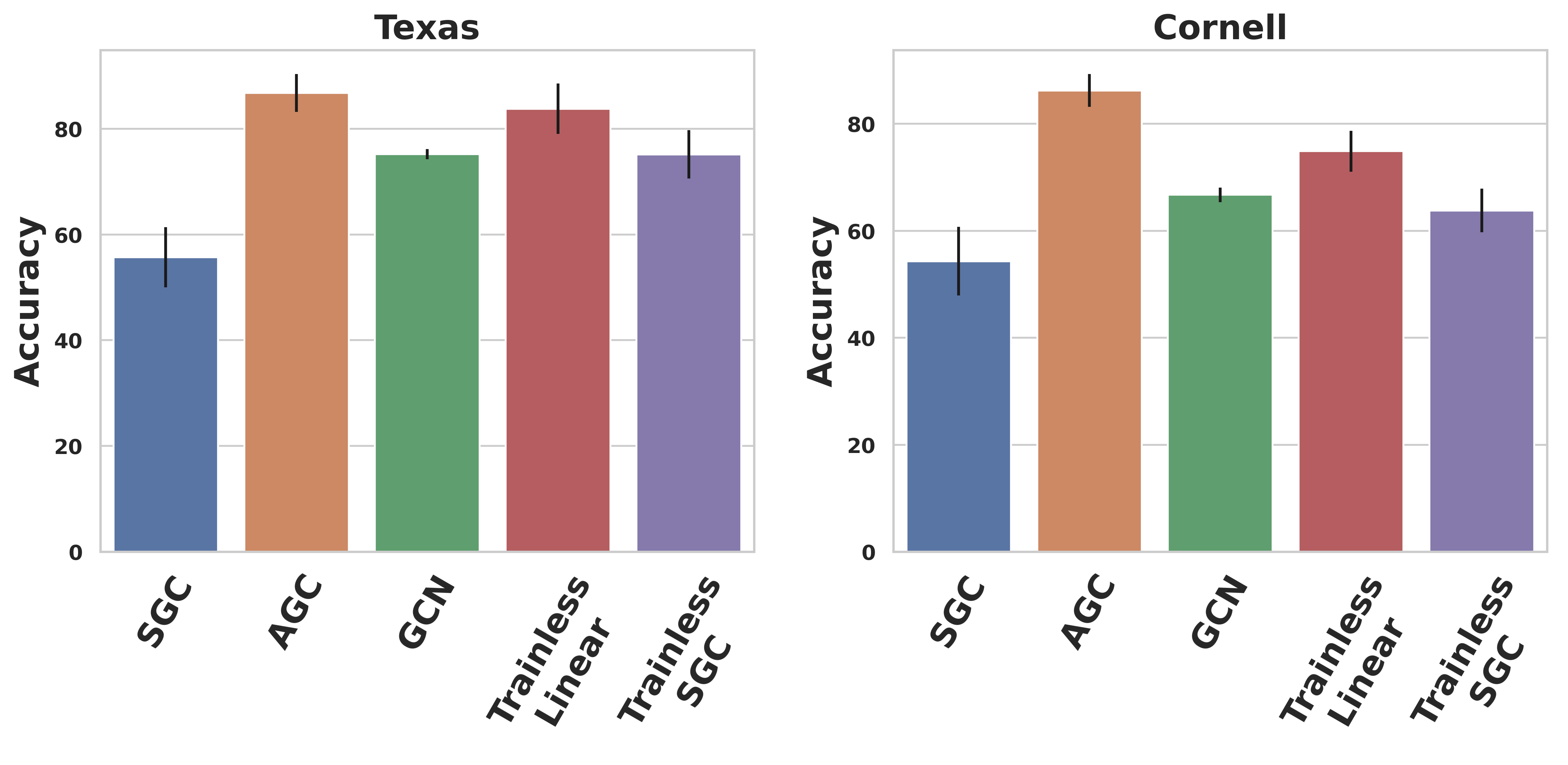}
    }
    \caption{Performance of our methods on heterophilous graphs.\vspace{-2mm}}
    \label{fig:heter}
    \vspace{-2mm}
\end{figure}
Typical GNNs like \textbf{GCN} and \textbf{SGC} generally assume graph homophily, where nodes predominantly link to similar nodes~\cite{nt_revisiting_2019}. However, this isn't always the case. To assess our trainless models' effectiveness on heterophilous graphs, we experiment with two such graphs: Texas and Cornell~\cite{craven_learning_2000}. We include \textbf{AGC}~\cite{chanpuriya_simplified_2022}, a baseline model designed for heterophilous graphs, for comparison. As Figure \ref{fig:heter} illustrates, our model, \our, not only surpasses \textbf{GCN} and \textbf{SGC} but also delivers performance on par with \textbf{AGC}. This demonstrates \our's adaptability to both homo/heterophilous graph structures.
\vspace{-3mm}
\subsection{Training efficiency}
Our experiments on training efficiency show that our trainless methods are markedly faster than traditional gradient descent optimization. Owing to its one-step computation, \our are up to two orders of magnitude quicker than conventionally trained GNNs, including GCN and SGC. Detailed comparisons are provided in Figure \ref{fig:speed} in the Appendix.

%% file: body/conclusion.tex
\section{Conclusion}
In this study, we ventured into an alternative approach to fitting the GNN model for addressing the semi-supervised node classification problem on TAG, bypassing the traditional gradient descent training process. We analyze the distinctive challenges inherent to semi-supervised node classification and investigate the training dynamics of GNNs on text-attributed graphs. Subsequently, we introduce \our, a novel method capable of fitting a linear GNN without resorting to the gradient descent training procedure. Our comprehensive experimental evaluations show that our trainless models can either align with or even outperform their traditionally trained counterparts.

%% file: body/appendix.tex
\begin{figure*}[h]
    \centering
    \includegraphics[width=1.0\textwidth]{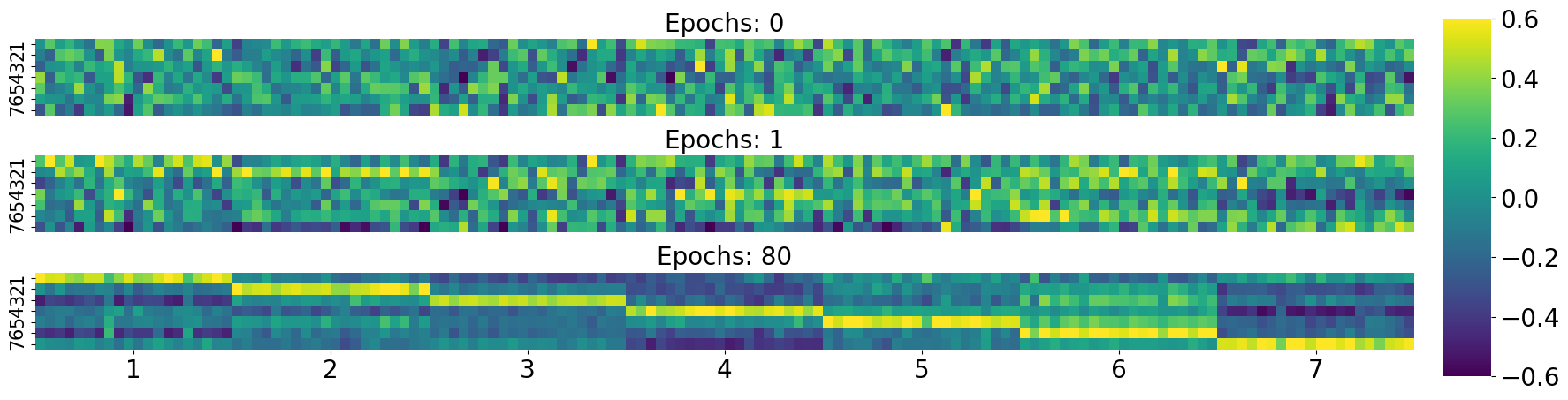}
    \caption{Heatmap depicting the evolution of inner products between node representations from GCN's first layer and the second layer's weight vectors across various training epochs on the Cora dataset.}
    \label{fig:gcn_training}
\vspace{-3mm}
\end{figure*}

\section{More strategical design}
While the weight matrix obtained by \Eqref{eq:neg_mp} is sufficient for a predictive \our, we further enhance the performance by introducing two more strategical design for our method.

\paragraph{Utilizing Propagated Node Attributes.} In updating the representation of virtual label nodes during message passing in \Eqref{eq:neg_mp}, nodes in the training set are initially associated with their original node attributes $\mathbf{X}_{\mathcal{L}}$. An alternative approach is to initialize the labeled nodes with a smoothed version of the node attributes. To execute this, prior to integrating the virtual label nodes into the graph, we conduct $l$ rounds of message passing as defined by $\text{AGG}(\cdot)$. This step refines each labeled node's representation over the graph structure. Post this refinement, we introduce the virtual label nodes into the graph and proceed as earlier, but now associating labeled nodes with the propagated node attributes. Specifically, the weight matrix $ \mathbf{W} $ is now obtained from \Eqref{eq:neg_mp} as:
\begin{align}
\label{eq:prop}
    \mathbf{W}^\top = (\mathbf{B}_{\mathcal{L}}-\frac{\omega}{C}\mathbf{1})^\top \mathbf{H}^{(l)}_{\mathcal{L}},
\end{align}
where $ \mathbf{H}^{(l)}_{\mathcal{L}} $ denotes the smoothed node representation of labeled nodes. This adjustment is beneficial when the inner product within the same class nodes is not significant. The propagated node attributes can help keep the node attributes closer to the corresponding class's trainless weight vector. It's worth noting that any GNN can implement the message passing function $ \text{AGG}(\cdot) $. In practice, the message passing settings of GCN/SGC are chosen for use.

\paragraph{Weighted message passing.} The method used to obtain the weight matrix by propagating node attributes from the same class as in \Eqref{eq:prop} treats each node equally, without considering their local structural information. To integrate this structural information into the weight matrix calculation, we introduce a weighted message passing technique. This approach is inspired by the Adamic Adar index (AA) \cite{adamic_friends_2003} and Resource Allocation (RA) \cite{zhou_predicting_2009}, utilized in link prediction. In this context, messages from nodes with fewer neighbors are given more weight compared to messages from hub nodes. On the contrary, the unweighted version can be seen as an approach akin to Common Neighbor (CN) \cite{liben-nowell_link_2003}. The formal expression for message passing is then reformulated as:
\begin{align}
\label{eq:weighted_mp}
    \mathbf{W}^\top = \mathbf{R}_{\mathcal{L}}(\mathbf{B}_{\mathcal{L}}-\frac{\omega}{C}\mathbf{1})^\top  \mathbf{H}^{(l)}_{\mathcal{L}},
\end{align}
where $ \mathbf{R}_{\mathcal{L}} $ is the degree normalization matrix specific to the labeled nodes. To compute $ \mathbf{R}_{\mathcal{L}} $, we initially determine the normalization matrix $ \mathbf{R} $ for the entire node set, which encompasses both $ \mathcal{L} $ and $ \mathcal{U} $. Based on the degree matrix $ \mathbf{D} $, for AA, we have $ \mathbf{R} = (\log \mathbf{D})^{-1} $ and for RA, $ \mathbf{R} = \mathbf{D}^{-1} $. For the CN-type message passing, we simply set $ \mathbf{R} = \mathbf{I} $. $ \mathbf{R}_{\mathcal{L}} $ is then acquired by extracting the corresponding portions from $ \mathbf{R} $ for labeled nodes.

\section{Supplementary experiments}


\begin{figure*}[h!]
    \centering
    \resizebox{\textwidth}{!}{%
    \includegraphics[width=1.0\linewidth]{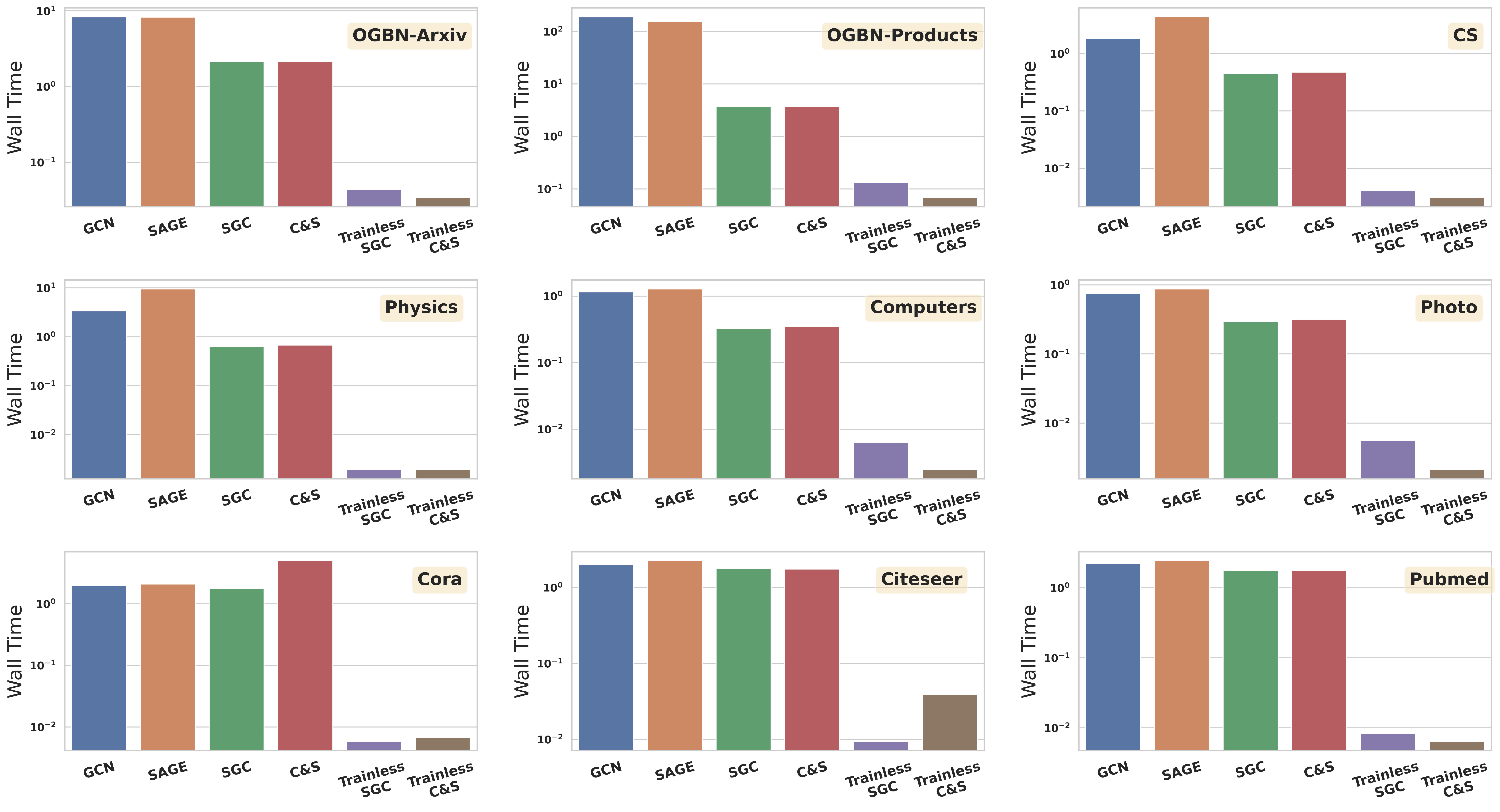}
    }
    \caption{Training efficiency of \our compared to the traditional gradient descent optimizations.}
    \label{fig:speed}
\end{figure*}

\subsection{Statistics of benchmark datasets}
\begin{table*}[h]
\caption{The statistics of the benchmark datasets.}
        \begin{center}
        \resizebox{\textwidth}{!}{
        \begin{tabular}{lcccccc}
        \toprule
            \textbf{Dataset} & \textbf{\#Nodes} & \textbf{\#Edges} & \textbf{\#Classes} & \textbf{Attr. Dimension} & \textbf{Text Encoding} & \textbf{\#Training/\#Validation/\#Testing} \\
            \midrule
            \textbf{Cora} & 2708 & 10556 & 7 & 1433 & BOW & 140/500/1000\\
            \midrule

            \textbf{Citeseer} & 3327 & 9104 & 6 & 3703 & BOW & 120/500/1000\\
            \midrule

            \textbf{Pubmed} & 19717 & 88648 & 3 & 500 & TF-IDF & 60/500/1000\\
            \midrule

            \textbf{CS} & 18333 & 163788 & 15 & 6805 & BOW & 300/450/17583\\
            \midrule

            \textbf{Physics} & 34493 & 495924 & 5 & 8415 & BOW & 100/150/34243\\
            \midrule

            \textbf{Computers} & 13752 & 491722 & 10 & 767 & BOW & 200/300/13252\\
            \midrule

            \textbf{Photo} & 7650 & 238162 & 8 & 745 & BOW & 160/240/7250\\
            \midrule

            \textbf{OGBN-Products} & 2449029 & 123718152 & 47 & 100 & BOW+PCA & 196615/39323/2213091\\
            \midrule

            \textbf{OGBN-Arxiv} & 169343 & 2315598 & 40 & 128 & Average of word embedding & 90941/29799/48603\\
        \bottomrule
        \end{tabular}
        }
        \end{center}
        \label{tab:stats}
\end{table*}

We show the statistics of the benchmark datasets in \autoref{tab:stats}. The data reveals that the three Planetoid datasets (Cora, Citeseer, and Pubmed) along with the two Coauthor datasets (CS and Physics) exhibit clear characteristics of over-parameterization and quasi-orthogonality. These traits contribute to \our achieving superior performance even when compared to trained models. Conversely, the Amazon co-purchase networks, despite showing over-parameterization, exhibit weaker quasi-orthogonality (See \Figref{fig:heatmap}). For the other two OGB datasets (OGBN-Arxiv and OGBN-Products), the abundance of labeled nodes in the training sets alleviates the semi-supervised node classification task for the trained models.

\subsection{Baseline model details}
\paragraph{Baseline method \textbf{C\&S}}
Recall that the C\&S approach~\cite{huang_combining_2020} commences training with an MLP solely on node attributes, disregarding the graph structure. It then propagates the computed logits through the graph to refine predictions based on the graph structure, as formulated:
\begin{align}
\mathbf{\hat{Z}} = \text{MLP}(\mathbf{X}), \mathbf{Z}= \text{C\&S}(\mathbf{\hat{Z}},\mathbf{A}),
\end{align}
where $\mathbf{\hat{Z}}$ represents the logits for the node classification tasks. The logit propagation process in $\text{C\&S}(\cdot)$ encompasses a two-step sequence, namely the $\text{Correct}(\cdot)$ step and $\text{Smooth}(\cdot)$ step:
\begin{align}
    \text{C\&S} = \text{Smooth} \circ \text{Correct} (\mathbf{\hat{Z}},\mathbf{A},\mathbf{B}),
\end{align}
where $\mathbf{B}$ denotes the one-hot encoding of the labels.

The $\text{Correct}(\cdot)$ step is implemented as:
\begin{align}
\mathbf{E}^{(\ell)} &= \alpha_1 \mathbf{D}^{-1/2}\mathbf{A} \mathbf{D}^{-1/2} \mathbf{E}^{(\ell - 1)} + (1 - \alpha_1) \mathbf{E}^{(\ell - 1)}\\
\mathbf{{Z}^\prime} &= \mathbf{Z} + \gamma \cdot \mathbf{E}^{(L_1)},
\end{align}
where $\mathbf{E}^{(\ell)} \in \mathbb{R}^{n \times C}$ symbolizes the error matrix and $\gamma$ represents the scaling factor. The error matrix is defined as follows:
\begin{align}
\begin{split}
\mathbf{E}^{(0)}_i &= \begin{cases}
    \mathbf{B}_i - \mathbf{\hat{Z}}_i, & \text{if } v_i \in \mathcal{L} \text{,}\\
    \mathbf{0}, & \text{else}.
\end{cases}
\end{split}
\end{align}

Subsequently, the $\text{Correct}(\cdot)$ step is defined as:
\begin{align}
    \begin{split}
    \mathbf{Z}^{(0)}_i &= 
    \begin{cases}
    \mathbf{B}_i, & \text{if } v_i \in \mathcal{L} \text{,}\\
    \mathbf{{Z}^\prime}_i, & \text{else}
    \end{cases}
    \end{split}
    \\
    \mathbf{Z}^{(\ell)} &= \alpha_2 \mathbf{D}^{-1/2}\mathbf{A}
\mathbf{D}^{-1/2} \mathbf{Z}^{(\ell - 1)} +
(1 - \alpha_2) \mathbf{Z}^{(\ell - 1)}\\
\mathbf{Z} &= \mathbf{\hat{Z}}^{(L_2)}.
\end{align}

For a fair comparison, we employ a linear classifier as the base predictor for the C\&S model in the experiment section's baseline model. This choice is motivated by the fact that the trainless version of the C\&S model is also implemented with solely a linear classifier serving as the predictor.

\subsection{Software and hardware details}
We implement~\our in Pytorch Geometric framework~\cite{fey_fast_2019}. We run our experiments on a Linux system equipped with an NVIDIA A100 GPU.

\subsection{Hyperparameter selections}
Our method only has two hyperparameters governing the fitted weight matrix, which makes it easy to find the optimal hyperparameter setting. The two hyperparameters are the edge weight $\omega$ and the degree normalization matrix $ \mathbf{R} $. Thus, we tune $\omega$ in $[-1,0,0.001,0.01,0.1,1]$. For $ \mathbf{R} $, we select the message passing types in \Eqref{eq:weighted_mp} between CN, AA and RA. For the baseline methods, we train for $100$ epochs and fine-tune the number of GNN layers, the learning rate, and the $l$-2 norm penalty. All the final result is selected based on the model's performance on the validation set.

\subsection{Parameter Sensitivity}
\begin{figure*}[h!]
    \centering
    \resizebox{\textwidth}{!}{%
    \includegraphics[width=1.0\linewidth]{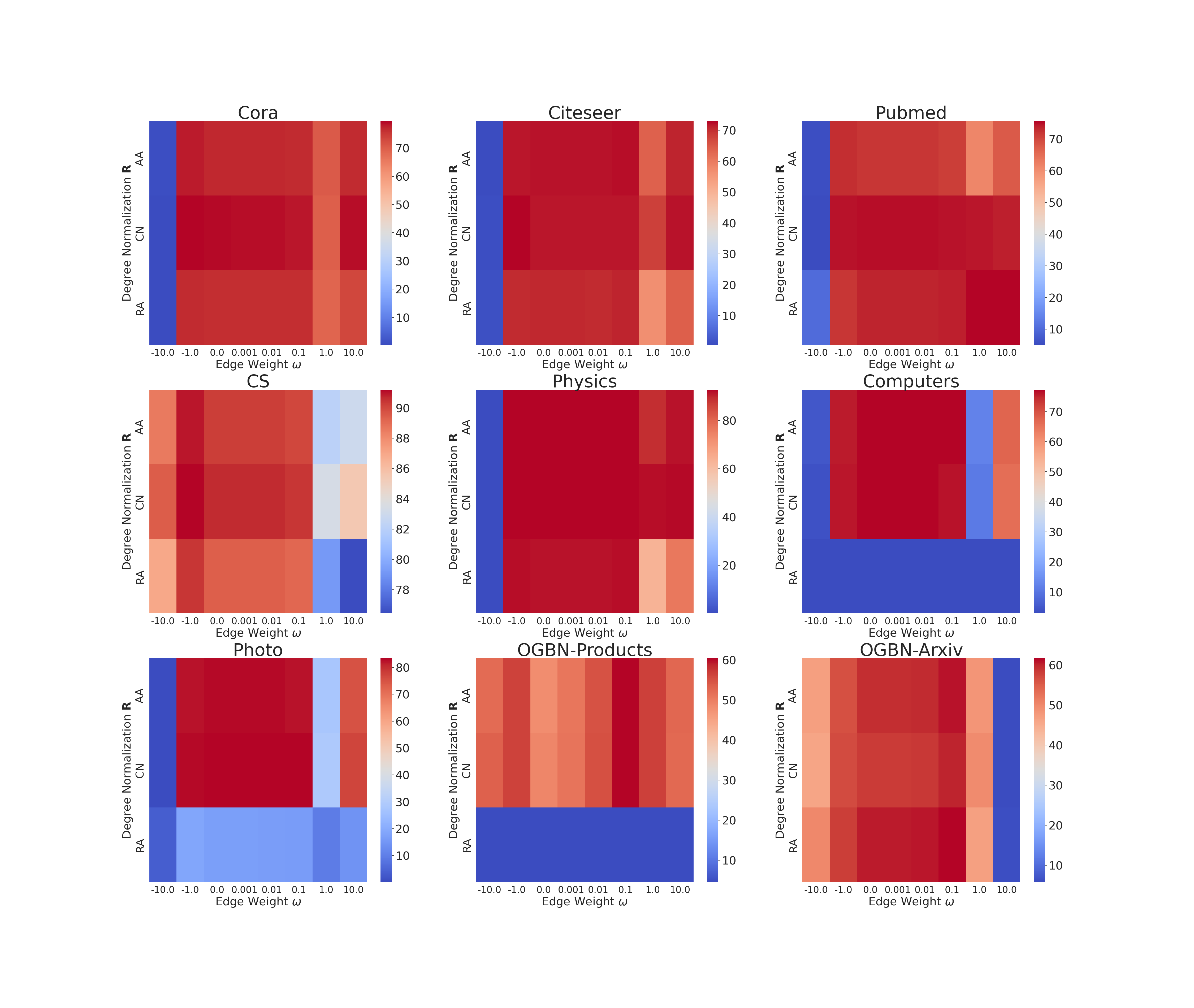}
    }
    \caption{Parameter sensitivity analysis for degree normalization matrix $\mathbf{R}$ and edge weight $\omega$ using the \textbf{Trainless SGC} model.}
    \label{fig:param_sen_sgc}
\end{figure*}

We further conduct a parameter sensitivity analysis to investigate the impact of the hyperparameters on \our. This analysis is performed using \textbf{Trainless SGC}, with the results shown in \Figref{fig:param_sen_sgc}. The findings underscore that the optimal degree normalization matrix, $\mathbf{R}$, may vary across different benchmarks due to their unique data attributes. For example, RA-type message passing excels on the Pubmed and OGBN-Arxiv datasets, while CN and AA-type message passing proves more robust on the remaining datasets. Concurrently, the edge weight, $\omega$, remains consistent across varied settings. However, a larger setting of $\omega$ ($\omega=1$) can adversely affect the performance of our trainless methods by excessively penalizing the node attributes from other classes. Intriguingly, even with a slightly negative value of $\omega$, our trainless model retains its efficacy. In such scenarios, the weight vector consists of not only the node attributes from the corresponding class but also from other classes. Yet, a continual reduction in $\omega$ leads to a marked performance decline, indicating that the weight vectors could become indistinguishable under such conditions.

\subsection{Experimental details on heterophilous graphs}
On heterophilous graphs, we take the same experimental protocol as~\cite{chanpuriya_simplified_2022} to partition the datasets into $10$ different splits. The splits are predefined as~\cite{craven_learning_2000}. We run the same set of hyperparameter selections as the homophilous graph experiments.

\begin{figure*}[h]
    \centering
    \resizebox{\textwidth}{!}{%
    \includegraphics[width=1.0\linewidth]{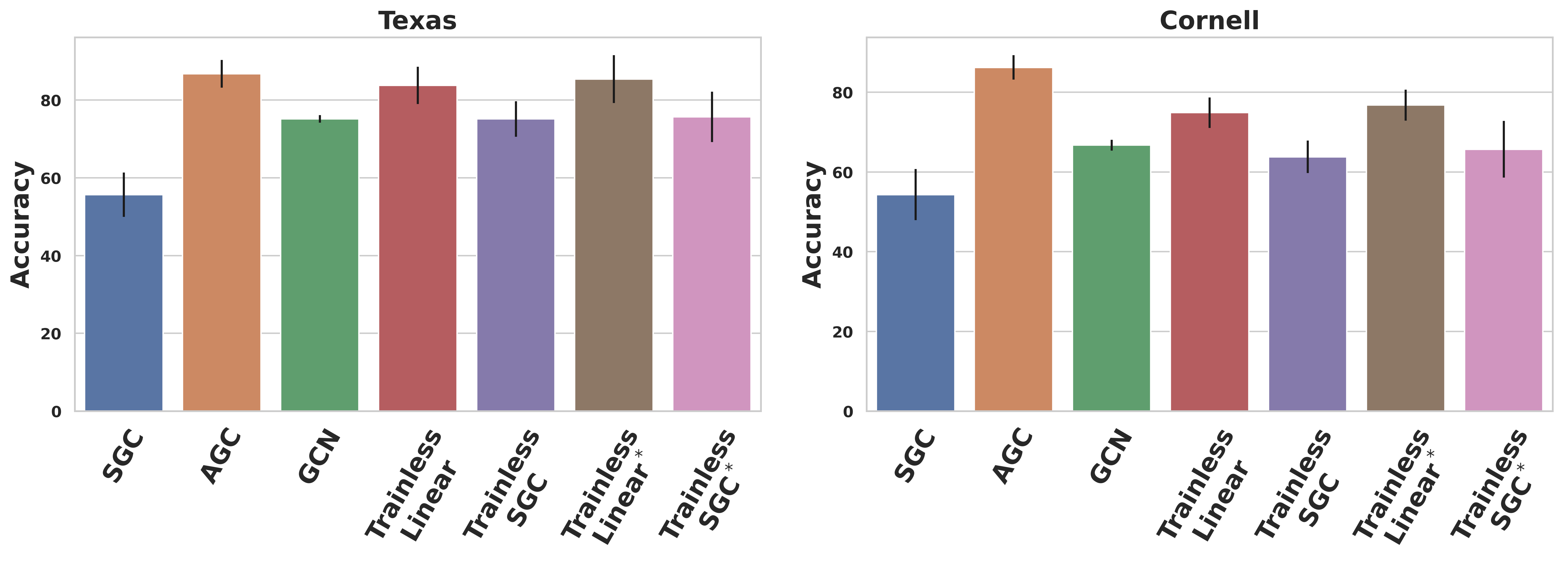}
    }
    \caption{Performance of our methods on heterophilous graphs. $^*$ indicates the models trained with training and validation labels.}
    \label{fig:heter_val}
\end{figure*}

We also conduct an experiment that fits the model including labels from validation sets. The results are shown in Figure \ref{fig:heter_val}. We can see that the benefits of bringing more validation labels are marginal to \our on the heterophilous graphs. 

\section{Limitations}
Despite the efficacy of our proposed method in tackling the semi-supervised node classification problem on text-attributed graphs, certain limitations prevail. Firstly, our method is confined to fitting a linear GNN model, which may limit the model's expressiveness. Secondly, the successful deployment of our method depends on specific data configurations, namely over-parameterization, potentially narrowing its applicability across a broader spectrum of cases.